\documentclass[letterpaper, 10 pt, conference]{ieeeconf}  

\IEEEoverridecommandlockouts                              

\usepackage{graphicx} 
\usepackage{algorithm}
\usepackage{algpseudocode}
\usepackage{bibentry}
\usepackage{amssymb}
\usepackage{amsmath}
\usepackage{multirow}
\usepackage{graphicx}
\usepackage{subcaption} 
\usepackage[table]{xcolor}  
\definecolor{waymolgray}{HTML}{F0F0F0} 
\usepackage{tabularx} 
\usepackage{booktabs} 
\usepackage{array}    
\usepackage{cleveref}
\usepackage{xcolor}
\usepackage{pifont}

\title{\LARGE \bf
DriveVer: Lightweight Trajectory Evaluator as Test-Time Verifier for Autonomous Driving
}

\author{
    Chong He\textsuperscript{1*},
    Yuechen Luo\textsuperscript{1*}, 
    Fang Li\textsuperscript{2}, 
    Shaoqing Xu\textsuperscript{2}, 
    Fuxi Wen\textsuperscript{1,\ding{41}}
\thanks{\textsuperscript{1}Tsinghua University, 
\textsuperscript{2}University of Macau}
\thanks{*Equal contribution.  \ding{41} Corresponding author.}
}

\begin{document}

\maketitle
\thispagestyle{empty}
\pagestyle{empty}

\begin{abstract}
End-to-end autonomous driving models often encounter performance bottlenecks, as training-time scaling leads to high computational costs and diminishing marginal returns. Existing planners typically adopt a one-shot generation paradigm, lacking secondary validation and active correction mechanisms to detect and revise suboptimal or unsafe trajectories during inference. To address this issue, we propose DriveVer, a lightweight, plug-and-play Test-Time Verifier that leverages the test-time scaling paradigm to enable autonomous driving systems to validate and refine trajectories without costly and heavy training. We construct a dedicated trajectory dataset based on the NAVSIM benchmark through condition-driven clustering and balanced sampling according to ego-vehicle states and navigation commands. Employing a dual-head architecture, DriveVer efficiently fuses candidate trajectories with multi-view visual representations and ego-vehicle kinematic features to simultaneously predict a safety confidence score and an absolute geometric refinement vector. Extensive experiments on the NAVSIM benchmark show that DriveVer significantly improves the performance of base planning models. Notably, as an extremely compact model with only 34M parameters, DriveVer introduces minimal computational overhead, achieving competitive results while maintaining real-time inference efficiency.
\end{abstract}

\section{Introduction}
In recent years, end-to-end autonomous driving has shown remarkable progress in handling complex urban driving scenarios~\cite{jiang2025survey,hu2025vision}. To improve robustness against long-tail and rare corner cases, current research primarily relies on training-time scaling, increasing model capacity and training data to enhance planning performance. However, this strategy incurs substantial computational cost while yielding diminishing performance gains as model scale continues to grow.

More importantly, most existing end-to-end planners adopt a one-shot generation paradigm~\cite{chen2024end,jiang2023vad,hu2023planning,adathinkdrive,zhang2026minddriver,lu2026xiaomionevlonesteplatent,chen2026vilta}, where the planning trajectory is generated once from sensor observations and directly executed by the downstream controller. Consequently, the planner cannot verify or revise its prediction during inference. When encountering complex or adversarial scenarios, suboptimal trajectory predictions cannot be effectively detected or corrected, limiting the safety and reliability of autonomous driving systems.

Recently, test-time scaling, also referred to as test-time compute, has emerged as a promising paradigm for improving model performance through additional inference-time computation (e.g., in large language models~\cite{zhang2025survey}). Rather than relying solely on larger models, it enables iterative verification and self-correction during inference. However, directly applying this paradigm to autonomous driving remains challenging due to stringent real-time constraints, as existing inference mechanisms often introduce computational overhead incompatible with the millisecond-level latency requirements of end-to-end planning~\cite{chen2024end}.

\begin{figure}[t!]
    \centering
    \includegraphics[width=\linewidth]{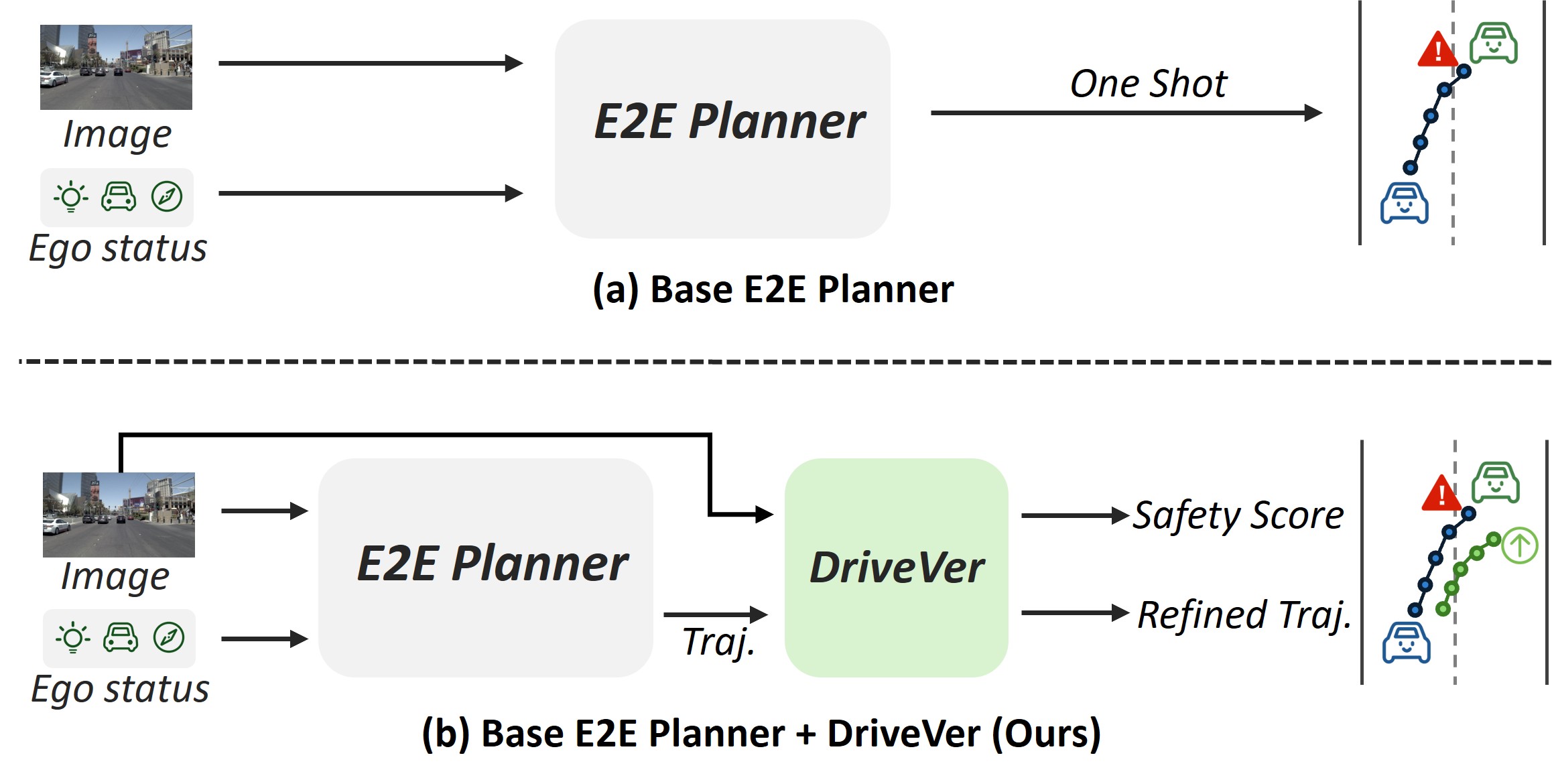}
    \caption{\textbf{Conceptual comparison of different planning paradigms.} (a) The base planner directly outputs a one-shot trajectory for execution, which can be unsafe. (b) Our proposed DriveVer refines this initial trajectory at test time, providing a safety score and a corrected trajectory.}
    \vspace{-5mm}
    \label{fig:intro}
\end{figure}

To address this gap and overcome the limitations of pure training-time scaling, we propose DriveVer, a lightweight post-processing framework for trajectory evaluation and refinement. As illustrated in Fig.~\ref{fig:intro}, we compare the conventional planning pipeline with our verification paradigm. Unlike existing end-to-end planners that directly execute a one-shot trajectory, potentially leading to unsafe behaviors in challenging scenarios (Fig.~\ref{fig:intro}(a)), DriveVer introduces an additional safety-critical verification stage at test time. As a plug-and-play Test-Time Verifier, DriveVer operates without modifying the architecture of the underlying planner. Instead, it takes the initial planned trajectory together with sensor observations and ego-state information to estimate trajectory quality and perform geometric refinement (Fig.~\ref{fig:intro}(b)), producing a refined trajectory with improved safety and reliability.

A key challenge in training such a verifier is the lack of dense supervision in existing autonomous driving datasets. To address this issue, we construct a dedicated trajectory verification and refinement dataset based on the NAVSIM benchmark. Specifically, human driving trajectories are grouped according to ego-vehicle kinematic states (e.g., speed and acceleration) and high-level navigation commands, followed by strict clustering and multiple rounds of sampling to generate diverse candidate trajectories for each driving scenario. By comparing each candidate trajectory with the corresponding human demonstration, we obtain accurate quality scores and geometric refinement targets, providing dense supervision for both trajectory evaluation and refinement.

DriveVer is designed as a lightweight and computationally efficient framework. It jointly encodes candidate trajectories, high-level navigation commands, ego-vehicle kinematic states (e.g., speed and acceleration), and compressed multi-view visual features through a cross-modal cross-attention mechanism. The network adopts a dual-head architecture, where one branch estimates trajectory quality and safety scores, while the other predicts geometric refinement vectors for trajectory correction. Extensive experiments on the NAVSIM benchmark demonstrate that, as a plug-and-play post-processing module, DriveVer consistently improves the performance of diverse end-to-end planners while introducing only minimal computational overhead.

In summary, the main contributions of this paper are as follows:

\begin{itemize}
    \item We propose a lightweight test-time scaling paradigm for autonomous driving: we design DriveVer, a plug-and-play Test-Time Verifier. It breaks the traditional paradigm of improving performance only through extensive training and demonstrates that low-cost trajectory scoring and refinement during inference can effectively tighten the system's safety upper bound.
    \item We construct a condition-driven trajectory verification dataset: based on the NAVSIM benchmark, we propose a trajectory clustering and sampling strategy based on ego-vehicle kinematic states (speed, acceleration) and navigation commands, constructing a dataset containing diverse candidate trajectories and their corresponding real quality scores for each scenario, filling the data gap in this field.
    \item We design an efficient cross-modal trajectory refinement architecture, DriveVer, that fuses visual representations and ego-vehicle dynamic features to achieve accurate scoring and geometric-level absolute refinement of initial trajectories. Experiments show that this module achieves highly competitive performance while maintaining extremely high inference efficiency.
\end{itemize}

\section{Related Work}
\subsection{End-to-end Autonomous Driving}
UniAD~\cite{hu2023planning} pioneers a fully differentiable end-to-end framework that unifies multiple perception tasks, demonstrating the feasibility of holistic end-to-end autonomous driving. Building upon this paradigm, VAD~\cite{jiang2023vad} improves computational efficiency by adopting a vectorized scene representation. Subsequent methods, including VADv2~\cite{chen2024vadv2} and Hydra-MDP~\cite{li2024hydra}, further advance the framework by introducing multi-modal planning based on rule-guided scoring and sampling from a fixed set of anchor trajectories. More recently, diffusion models have emerged as a promising paradigm for end-to-end autonomous driving, owing to their ability to capture complex multi-modal distributions in high-dimensional action spaces through an iterative denoising process. DiffusionDrive~\cite{liao2025diffusiondrive} mitigates the mode collapse issue by introducing an anchor-based truncated denoising strategy and significantly improves inference efficiency by reducing the denoising process to only two steps. These advances suggest that diffusion-based generative models offer a promising direction for end-to-end autonomous driving by balancing trajectory quality and inference efficiency.

\subsection{VLA for Autonomous Driving}
Vision-language models (VLMs) have attracted increasing attention in autonomous driving by enabling unified perception, planning, and decision-making through the integration of visual and textual information. Existing methods can be broadly categorized into two paradigms. The first focuses on scene understanding and high-level reasoning~\cite{jiang2024senna,jiang2025alphadrive,mtrdrive}, where methods such as Senna~\cite{jiang2024senna} generate meta-actions from sensory inputs for downstream planners but have shown limited improvements in real-world driving performance. The second paradigm directly predicts driving trajectories from raw sensory inputs~\cite{hwang2024emma,xing2025openemma,adathinkdrive,zhou2025autovla,li2025recogdrive,luo2026unleashing,luo2026last}. A notable trend within this line of research is the incorporation of intermediate reasoning, such as Chain-of-Thought (CoT), to improve model interpretability and planning accuracy. Methods including EMMA~\cite{hwang2024emma}, ReasonPlan~\cite{liu2025reasonplan}, and Sce2DriveX~\cite{zhao2025sce2drivex} demonstrate that domain-specific reasoning effectively enhances trajectory prediction performance.

\begin{figure*}[t!]
    \centering
    \includegraphics[width=\linewidth]{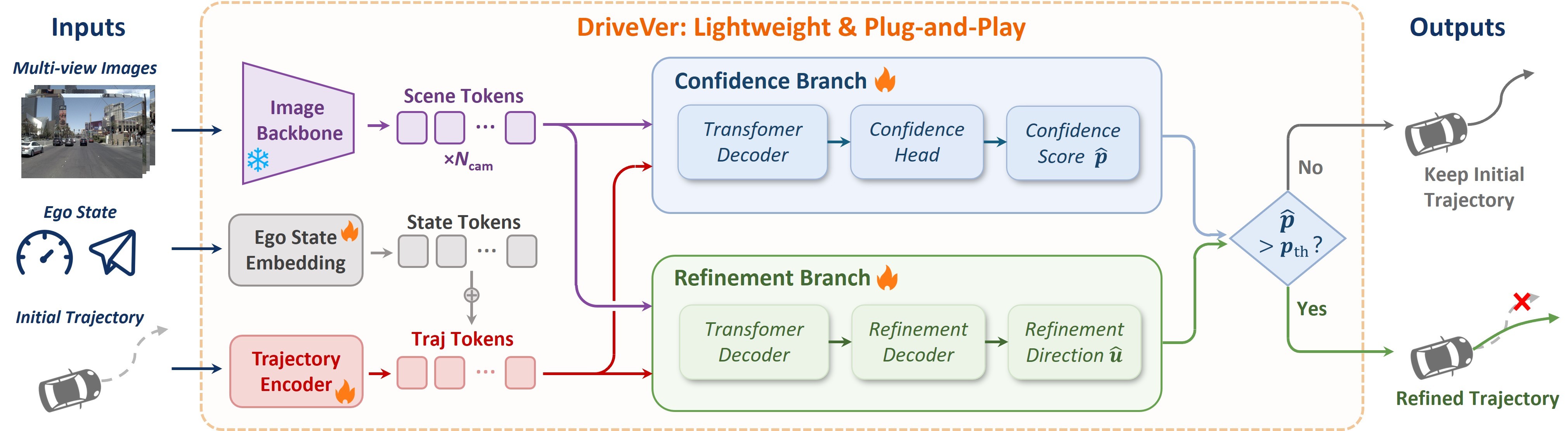}
    \caption{\textbf{DriveVer architecture.} DriveVer employs a dual-head output architecture for trajectory evaluation and optimization, in which the confidence branch outputs a scalar confidence score to determine the need for intervention, and the refinement branch generates a geometric refinement direction. During inference, once the predicted confidence score exceeds a predefined safety threshold, DriveVer will execute the refinement process to optimize the initial trajectory.}
    \vspace{-5mm}
    \label{fig:main}
\end{figure*}

\subsection{Trajectory Evaluation and Refinement} 
Trajectory evaluation plays a crucial role in autonomous driving by assessing candidate trajectories and filtering unsafe predictions. Existing methods can be broadly categorized into model-free approaches, which assign heuristic scores using rule-based metrics or explicit perception annotations, and model-based approaches, which leverage learned environment dynamics or intermediate representations, such as BEV features in WOTE~\cite{li2025end} and perception caches in Hydra-MDP~\cite{li2024hydra}. Beyond evaluation, improving safety also requires trajectory refinement. Most multi-modal end-to-end planners generate a large set of anchor trajectories and use the evaluation module solely to select the highest-scoring candidate, while conventional refinement methods instead rely on computationally intensive optimization-based smoothers or rule-based safety filters to post-process the selected trajectory.

Despite the recent progress in learning-based evaluators such as DriveReward~\cite{chen2026drivereward}, existing evaluation and refinement paradigms exhibit two key limitations. First, their dependence on explicit perception annotations or complex environment dynamics makes them computationally expensive and difficult to deploy as lightweight post-hoc modules. Second, framing trajectory evaluation purely as a selection problem over fixed anchor trajectories overlooks the opportunity for direct geometric refinement. As a result, current one-shot planning frameworks still lack an efficient and unified mechanism that can simultaneously estimate trajectory quality and generate corrective geometric refinements for suboptimal initial trajectories.

\section{Method}
This section presents DriveVer, a lightweight test-time trajectory verifier~(Fig.~\ref{fig:main}). Following the problem formulation, we detail its three core components: (1) the construction of a balanced trajectory dataset for training; (2) a dual-head model architecture for simultaneous trajectory evaluation and geometric refinement; and (3) the conditional test-time inference strategy.

\subsection{Preliminaries}

In this section, we formally define the problem setup and compare the traditional one-shot generation paradigm with our proposed test-time refinement framework.

\noindent\textbf{Base E2E Planner Formulation:}
In the traditional one-shot generation paradigm, at any given time step $t$, the base end-to-end planner receives a multimodal observation tuple $\mathbf{O}_t = (\mathbf{I}_t, s_{\text{ego}}, c)$. Specifically:
\begin{itemize}
    \item $\mathbf{I}_t$ represents the multi-view camera images (e.g., front, front-left, front-right, and back cameras).
    \item $s_{\text{ego}} \in \mathbb{R}^{d_s}$ is the kinematic state vector for the ego-vehicle, which explicitly includes real-time speed and acceleration.
    \item $c$ denotes the high-level navigation commands guiding the vehicle's route.
\end{itemize}

Given these inputs, the base planner directly generates a one-shot initial trajectory $\tau_{\text{init}} = \{(x_t,y_t,\theta_t)\}_{t=1}^T \in \mathbb{R}^{T \times 3}$ representing the planned future waypoints, where $T$ denotes the planning horizon, and $(x_t,y_t,\theta_t)$ is the location of each waypoint at time $t$ in the current ego-vehicle coordinate system. This trajectory is directly passed to the control layer for execution without any subsequent evaluation or correction.

\noindent\textbf{DriveVer Formulation:}
To address the lack of secondary validation and active correction, DriveVer is introduced as a plug-and-play Test-Time Verifier directly after the base planner. The input to DriveVer is an extended tuple $\mathbf{Q}_t = (\mathbf{I}_t, s_{\text{ego}}, c, \tau_{\text{init}})$, which uniquely incorporates the generated initial trajectory $\tau_{\text{init}}$ alongside the original sensor and state inputs.
DriveVer utilizes a dual-head output architecture to evaluate and optimize the given trajectory:
\begin{itemize}
    \item \textbf{Confidence Branch Output ($\hat{p}$):} A scalar safety score representing the refinement confidence, predicting the quality of the initial trajectory and determining whether it requires intervention.
    \item \textbf{Refinement Branch Output ($\hat{u}$):} An absolute geometric refinement direction vector for the trajectory.
\end{itemize}

During inference, if the predicted confidence score $\hat{p}$ exceeds a predefined safety threshold, DriveVer triggers the refinement process and optimizes the initial trajectory $\tau_{\text{init}}$, and the refined trajectory $\tau_{\text{refined}}$ is submitted as the final output for execution. Details will be discussed in the subsequent sections. 

\subsection{Data Collection}

High-quality training data is critical to enable the test-time verifier to reliably evaluate trajectories from the base planner and perform accurate trajectory refinement in real-world driving scenarios. Existing datasets often suffer from the issues of excessively optimal positive samples and sparse negative samples, making it difficult for the model to learn a fine-grained evaluation boundary. To this end, we construct a dedicated dataset for trajectory verification and refinement. Leveraging the NAVSIM~\cite{dauner2024navsim} dataset, we first construct a high-dimensional library of human driving primitives by performing unsupervised clustering and multi-stage sampling based on ego-vehicle kinematic manifolds (e.g., velocity, acceleration) and high-level navigational commands. For any given scenario, a candidate trajectory set $\{\tau_i\}_{i=1}^{N_c}$ is retrieved by matching the current state-space constraints within the cluster.

To endow the test-time verifier with the capacity to discriminate among trajectory quality and perform adaptive refinement, we propose a balanced trajectory construction pipeline to mitigate bias caused by an imbalanced sample distribution. Specifically, we first calculate the Predictive Driver Model Score (PDMS) for each candidate trajectory in $\{\tau_i\}_{i=1}^{N_c}$ as its quantitative quality metric. Based on the score distribution, we partition all trajectories into high-quality and low-quality subsets using a threshold of 0.5. To ensure the model learns a well-calibrated decision boundary, we perform class-balanced sampling by extracting $N_{half}$ trajectories from each subset, forming a representative and unbiased trajectory pool. We further augment this pool with the corresponding human expert trajectory $\tau_{\text{gt}}$ as the ground-truth reference, resulting in a unified trajectory set with $N_d=2N_{half}+1$ samples. By applying this pipeline to $N_{scene}$ diverse driving scenarios sampled from the NAVSIM dataset, we construct a dedicated, high-quality benchmark for trajectory quality assessment and adaptive refinement.

The merit of the proposed dataset resides in two aspects. First, by balancing the number of high-quality and low-quality trajectory samples within each scenario, DriveVer effectively captures the representative characteristics of both trajectory categories, thereby facilitating accurate trajectory evaluation. Second, equipped with guidance from expert trajectories, DriveVer can further infer reasonable and reliable refinement directions for defective trajectories.

\subsection{DriveVer Model Architecture}

\begin{figure}[t!]
    \centering
    \includegraphics[width=0.8\linewidth]{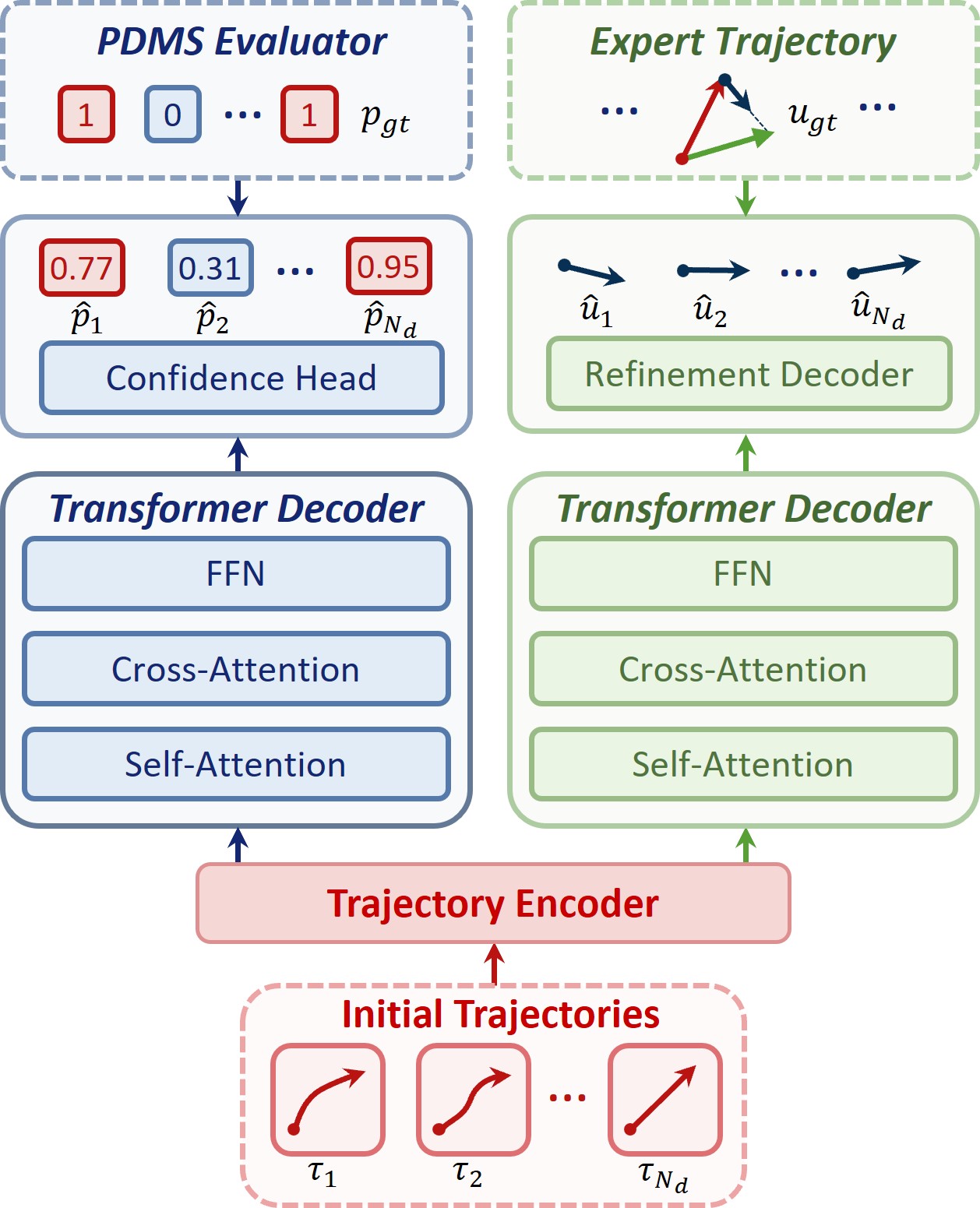}
    \caption{The architecture of the Transformer Decoder used in the confidence and refinement branches.}
    \vspace{-5mm}
    \label{fig:transformer}
\end{figure}
The core design concept of DriveVer is lightweight and highly efficient, serving as a plug-and-play module to assist the base planner. The overall model architecture is shown in Fig.~\ref{fig:main}. 
DriveVer takes multi-view camera images, ego-vehicle's kinematic states, high-level navigation commands, and initial trajectories as inputs. The image backbone in DriveVer is adapted from DrivoR~\cite{kirby2026driving}, which employs LoRA to enable parameter-efficient fine-tuning of the DINOv2 pre-trained Vision Transformer. It compresses perceptual information by assigning $R$ register tokens to each of the $N_{cam}$ cameras, and finally concatenates all register tokens to form $N_{cam} \times R$ scene tokens.

The initial trajectory is encoded by a trajectory encoder and fused with the ego-vehicle’s kinematic state and navigation command to form trajectory tokens. Subsequently, the trajectory tokens and scene tokens are fed into the trajectory refinement branch and the confidence branch, which predict the geometric refinement direction and the corresponding refinement confidence, respectively.

\subsubsection{Refinement Branch}

Given the initial trajectory and scene information, this branch estimates the potential refinement direction for the trajectory using a Transformer Decoder (Fig.~\ref{fig:transformer}) and a dedicated Refinement Decoder. During the training phase, we concatenate the trajectory tokens of the $N_d$ trajectories constructed in Sec. III-B as input to the Transformer Decoder. By performing cross-attention with the scene tokens, the refinement branch learns the underlying driving rules of the current scenario. The output of the Transformer Decoder is then fed into the Refinement Decoder to predict the refinement direction $\hat{u}$ for each trajectory.

The supervision signal for the refinement direction is obtained by comparing the initial trajectory with the human expert trajectory. Specifically, after flattening the initial trajectory and the human expert trajectory $\bar \tau_{\text{init}}, \bar \tau_{\text{gt}} \in \mathbb{R}^{3T \times 1}$, the normalized direction vector $u_{\text{gt}} \in \mathbb{R}^{3T \times 1}$ can be calculated as:

\begin{equation}
    u_{\text{gt}} = \frac{\bar \tau_{\text{gt}} - \bar \tau_{\text{init}}}{\|\bar \tau_{\text{gt}} - \bar \tau_{\text{init}}\|_2}
\end{equation}
where $\| \cdot \|_2$ is the Euclidean norm.
The training process minimizes the misalignment of cosine similarity between the predicted direction and the ground-truth direction:
\begin{equation}
\mathcal{L}_{dir} = \mathbb{E} \left[1-\langle \hat{u}, u_{\text{gt}} \rangle \right]
\end{equation}
where $\mathbb{E}$ is the expectation operator and $\langle \cdot, \cdot\rangle$ denotes the inner product of vectors.

The supervision of the refinement direction enables the model to focus more on the overall optimization trend of trajectories, prevents overfitting, and improves prediction robustness. The advantages of this formulation are also validated in subsequent ablation studies.

\subsubsection{Confidence Branch}

In addition to refining low-score trajectories, DriveVer should also be capable of retaining high-score trajectories, which requires the model to accurately assess the quality of the initial trajectory in the corresponding scenario and then decide whether to refine it. The confidence branch also comprises a Transformer Decoder and a Confidence Head. Similar to the refinement branch, the confidence branch learns to evaluate the quality of a trajectory under the given driving scenario by performing cross-attention with the scene tokens in the Transformer Decoder and then outputs a confidence score $\hat{p}$ for refinement via the Confidence Head.

The supervision signal of this branch is obtained as follows: we set the refinement confidence $p_{gt}$ of trajectories whose initial trajectory score exceeds the average score of human expert trajectories (94.8 in NAVSIM) to 0, indicating that the initial trajectory is sufficiently high-quality and no refinement is needed; the confidence of the other trajectories is set to 1. The binary
cross entropy (BCE) is selected as the loss function:
\begin{equation}
\mathcal{L}_{conf} = \mathbb{E} \left[\text{BCE}(\hat{p}, p_{gt}) \right]
\end{equation}

The total loss function uses the weighted sum of $\mathcal{L}_{dir}$ and $\mathcal{L}_{conf}$ :

\begin{equation}
    \mathcal{L} = \lambda \cdot \mathcal{L}_{dir} + (1-\lambda) \cdot \mathcal{L}_{conf}
\end{equation}

\subsection{Inference}

In the inference phase, DriveVer accepts sensor and state inputs, as well as the initial trajectory generated by any base planner. It first computes the confidence $\hat{p}$ to refine the current trajectory via the confidence branch. Only when the confidence exceeds the predefined threshold $p_{th}$ will it further output the refinement direction $\hat{u}$ through the trajectory refinement branch. The refined trajectory $\tau_{\text{refined}}$ is calculated as follows:
\begin{equation}
\tau_{\text{refined}} = \tau_{\text{init}} + \alpha \cdot \hat{u}
\end{equation}
where $\alpha$ denotes the magnitude of refinement. It enables DriveVer to adaptively control the refinement strength based on the performance of the base planner: for a base planner with satisfactory performance, only minor refinement is applied; otherwise, a larger correction amplitude is required. 

\section{Experiment}
\subsection{Experiment Setup}
\subsubsection{Dataset}
NAVSIM~\cite{dauner2024navsim} is a dataset derived from the real-world nuPlan~\cite{caesar2021nuplan} and is a subset of OpenScene~\cite{contributors2023openscene}. Sampled at 2 Hz, it emphasizes challenging scenarios involving changes in intention, where historical data cannot be extrapolated. In our experiments, we utilize the official \textit{navtrain} (85k samples) and \textit{navtest} (12k samples) splits for training and evaluation. The base planning model is trained on the 85k samples provided here. The dedicated training dataset for DriveVer is constructed by selecting 14,000 high-quality scenes from the \textit{navtrain} dataset and generating a set of 16 trajectories with scores spanning from low to high for each scene.

\begin{table*}[htbp] 
\centering 
\renewcommand{\arraystretch}{1.2} 
\setlength{\tabcolsep}{4.2mm}
\caption{Comparison with state-of-the-art methods on the NAVSIMv1 with PDMS.}
\label{table:navsimv1}
\begin{tabular}{l| c| c| ccccc| >{\columncolor{gray!30}}c} 
\toprule
\textbf{Method} & \textbf{Reference} & \textbf{Refine Policy} & \textbf{NC}$\uparrow$ & \textbf{DAC}$\uparrow$ & \textbf{TTC}$\uparrow$ & \textbf{CF}$\uparrow$ & \textbf{EP}$\uparrow$ & \textbf{PDMS}$\uparrow$ \\
\midrule
Constant Velocity &- & - & 68.0 & 57.8 & 50.0 & \textbf{100} & 19.4 & 20.6 \\
Ego Status MLP &- & - & 93.0 & 77.3 & 83.6 & \textbf{100} & 62.8 & 65.6 \\
\midrule
UniAD ~\cite{hu2023planning}  &CVPR 2023 &- & 97.8 & 91.9 & 92.9 & \textbf{100} & 78.8 & 83.4 \\
PARA-Drive ~\cite{weng2024drive} & CVPR 2024 & - & 97.9 & 92.4 & 93.0 & 99.8 & 79.3 & 84.0 \\
DiffusionDrive~\cite{liao2025diffusiondrive} & CVPR 2025 & - & 98.2 & 96.2 & 94.7 & \textbf{100} & 82.2 & 88.1 \\
WoTE~\cite{li2025end} & ICCV 2025 & - & 98.5 & 96.8 & 94.9 & 99.9 & 81.9 & 88.3 \\
Hydra-NeXt~\cite{li2025hydra} & ICCV 2025& - & 98.1 & 97.7 & 94.6 & \textbf{100} & 81.8 & 88.6\\
AutoVLA-3B~\cite{zhou2025autovla} & NeurIPS 2025& - & 98.4 & 95.6 & \textbf{100.0} & \textbf{100} & 81.9 & 89.1 \\
DriveVLA-W0-3B~\cite{li2025drivevla} & ICLR 2025& - & 98.7 & \textbf{99.1} & 95.3 & 99.3 & 83.3 & 90.3 \\
GoalFlow~\cite{xing2025goalflow} &CVPR 2025 & - & 98.4 & 98.3 & 94.6 & \textbf{100} & 85.0 & 90.3 \\
\midrule
\multirow{2}{*}{DiffusionDrive~\cite{liao2025diffusiondrive}} & \multirow{2}{*}{CVPR 2025} & Original & 98.2 & 96.2 & 94.7 & \textbf{100} & 82.2 & 88.1 \\
 & & with DriveVer & 98.3 & 96.8 & 94.2 & \textbf{100} & 84.1 & 89.0(\textcolor{orange!70!black}{+0.9}) \\
\midrule
\multirow{2}{*}{DrivoR~\cite{kirby2026driving}} & \multirow{2}{*}{CVPR 2026} & Original & \textbf{99.0} & 98.9 & 96.7 & \textbf{100} & 90.0 & 93.7 \\
 & & with DriveVer & \textbf{99.0} & 99.0 & 96.6 & \textbf{100} & \textbf{90.1} & \textbf{93.8}(\textcolor{orange!70!black}{+0.1}) \\
\midrule
\multirow{2}{*}{AdaThinkDrive~\cite{adathinkdrive}} & \multirow{2}{*}{ICRA 2026} & Original & 98.4 & 97.8 & 95.2 & \textbf{100} & 84.4 & 90.3 \\
 & & with DriveVer & 98.2 & 97.8 & 94.5 & \textbf{100} & 87.3 & 90.9(\textcolor{orange!70!black}{+0.6})  \\
\midrule
\multirow{2}{*}{ELF-VLA~\cite{luo2026unleashing}} & \multirow{2}{*}{CVPR 2026} & Original & 98.9 & 98.1 & 96.0 & \textbf{100} & 85.3 & 91.0 \\
 & & with DriveVer & 98.7 & 97.9 & 95.7 & \textbf{100} & 87.2 & 91.5(\textcolor{orange!70!black}{+0.5}) \\
\bottomrule
\end{tabular}
\end{table*}

\begin{table*}[t!]
\centering
\caption{Comparison with state-of-the-art methods on the NAVSIMv2 with EPDMS.}
\label{table:navsimv2}
\renewcommand{\arraystretch}{1.2} 
\resizebox{\textwidth}{!}{
\begin{tabular}{l|c|ccccc|cccc|>{\columncolor{gray!30}}c}
\toprule
\textbf{Method} & \textbf{Refine Policy} & \textbf{NC $\uparrow$} & \textbf{DAC $\uparrow$} & \textbf{DDC $\uparrow$} & \textbf{TLC $\uparrow$} & \textbf{EP $\uparrow$} & \textbf{TTC $\uparrow$} & \textbf{LK $\uparrow$} & \textbf{HC $\uparrow$} & \textbf{EC $\uparrow$} & \textbf{EPDMS $\uparrow$} \\
\midrule
Ego Status & - & 93.1 & 77.9 & 92.7 & 99.6 & 86.0 & 91.5 & 89.4 & 98.3 & 85.4 & 64.0 \\
TransFuser~\cite{chitta2022transfuser} & - & 96.9 & 89.9 & 97.8 & 99.7 & 87.1 & 95.4 & 92.7 & 98.3 & 87.2 & 76.7 \\
HydraMDP++~\cite{li2024hydra} & - & 97.2 & 97.5 & 99.4 & 99.6 & 83.1 & 96.5 & 94.4 & 98.2 & 70.9 & 81.4 \\
Recogdrive-8B~\cite{li2025recogdrive} & - & 98.3 & 95.2 & \textbf{99.5} & \textbf{99.8} & 87.1 & 97.5 & 96.6 & 98.3 & 86.5 & 83.6 \\
DriveVLA-W0-3B~\cite{li2025drivevla} & - & 98.5 & \textbf{99.1} & 98.0 & 99.7 & 86.4 & 98.1 & 93.2 & 97.9 & 58.9 & 86.1 \\
\midrule
 & Original & 98.2 & 95.9 & 99.4 & \textbf{99.8} & 87.5 & 97.3 & 96.8 & 98.3 & \textbf{87.7} & 84.5 \\
\multirow{-2}{*}{DiffusionDrive~\cite{liao2025diffusiondrive}} & with DriveVer & 98.3 & 96.9 & 99.4 & 99.7 & 88.8 & 97.4 & 96.7 & 98.3 & 86.7 & 85.5(\textcolor{orange!70!black}{+1.0}) \\
\midrule
 & Original & \textbf{99.0} & 98.9 & 98.8 & 99.6 & 92.1 & \textbf{98.5} & 95.1 & 97.7 & 70.1 & 86.5 \\
\multirow{-2}{*}{DrivoR~\cite{kirby2026driving}} & with DriveVer & 98.9 & 98.9 & 98.8 & 99.6 & \textbf{92.2} & \textbf{98.5} & 95.0 & 97.7 & 71.0 & 86.7(\textcolor{orange!70!black}{+0.2}) \\
\midrule
 & Original & 98.1 & 97.6 & 99.1 & 99.7 & 91.2 & 97.7 & 96.5 & 98.3 & 86.6 & 86.5 \\
\multirow{-2}{*}{AdaThinkDrive~\cite{adathinkdrive}} & with DriveVer & 98.2 & 97.8 & 99.1 & 99.7 & 91.4 & 97.7 & 96.4 & 98.3 & 86.5 & 86.7(\textcolor{orange!70!black}{+0.2}) \\
\midrule
 & Original & 98.9 & 98.1 & 99.4 & \textbf{99.8} & 88.5 & 98.4 & \textbf{96.9} & 98.3 & 87.2 & 87.1 \\
\multirow{-2}{*}{ELF-VLA~\cite{luo2026unleashing}} & with DriveVer & 98.7 & 97.9 & 99.3 & 99.7 & 90.6 & 98.3 & 96.6 & \textbf{98.4} & 86.5 & \textbf{87.2}(\textcolor{orange!70!black}{+0.1}) \\
\bottomrule
\end{tabular}
}
\end{table*}

\subsubsection{Metric}

As opposed to previous benchmarks such as nuScenes~\cite{qian2024nuscenes}, mostly formulating driving quality as a measure of similarity to the expert human trajectory, we utilize the Predictive Driver Model Score (PDMS) for \textbf{NAVSIMv1}~\cite{dauner2024navsim} and the Extended Predictive Driver Model Score (EPDMS) for \textbf{NAVSIMv2}~\cite{cao2025pseudo} as the closed-loop planning metrics.

For NAVSIMv1, PDMS integrates five sub-metrics: No At-Fault Collision (NC), Drivable Area Compliance (DAC), Time-to-Collision (TTC), Comfort (C), and Ego Progress (EP) to produce a comprehensive closed-loop planning score. Its calculation formula is defined as follows:
\begin{equation}
{PDMS} = NC \times DAC \times \left( \frac{5\times EP + 5\times TTC + 2\times C}{12} \right),
\end{equation}

For NAVSIMv2, the EPDMS metric includes several components categorized as penalties or weighted subscores. Its key metrics are No At-Fault Collision (NC), Drivable Area Compliance (DAC), Driving Direction Compliance (DDC), Traffic Light Compliance (TLC), Ego Progress (EP), Time to Collision (TTC), Lane Keeping (LK), History Comfort (HC), and Extended Comfort (EC). Its calculation formula is defined as follows:
\begin{equation}
\begin{split}
    EPDMS ={} & NC \times DAC \times DDC \times TLC \times \\
              & \left( \frac{5 EP + 2 LK + 2 HC + 5 TTC + 2 EC}{16} \right).
\end{split}
\end{equation}
\subsubsection{Implementation Details}
We adopted the visual encoder proposed by DrivoR, which takes RGB images from four cameras (front, front left, front right, and back) as input, assigns 16 learnable register tokens to each camera, and concatenates them to form 64 scene tokens. The parameters of the visual encoder are frozen during training. The decoders for both the refinement and confidence branches are 4-layer transformers with an inner dimension of 256. We adopt the AdamW optimizer with a learning rate of $10^{-4}$ and a cosine annealing learning rate schedule. Training is conducted for 30 epochs on our dedicated dataset, with the loss weight $\lambda$ fixed at 0.5. During inference, the confidence threshold for applying refinement is set to 0.75, and the refinement magnitude $\alpha$ is set to 0.5. 

\subsection{Main Results}
\subsubsection{Quantitative Evaluation}
We plug DriveVer into several end-to-end planners, including DiffusionDrive~\cite{liao2025diffusiondrive}, AdaThinkDrive~\cite{adathinkdrive}, ELF‑VLA~\cite{luo2026unleashing} and DrivoR~\cite{kirby2026driving}, to verify its effectiveness. Tab.~\ref{table:navsimv1} shows the performance comparison between the aforementioned base planners with and without DriveVer on the NAVSIMv1 benchmark. The results demonstrate that DriveVer can steadily improve the performance of the base planners by evaluating and refining their initial trajectories. Specifically, DiffusionDrive achieves a 0.9 improvement in PDMS after integrating DriveVer, yielding the most significant performance gain. Even for DrivoR, which achieves state-of-the-art performance among all base planners, a slight yet consistent PDMS improvement is observed after applying DriveVer, further demonstrating the effectiveness of DriveVer. On the NAVSIMv2 benchmark, DriveVer continues to demonstrate stable trajectory optimization capability. 

The results indicate that, by training on our constructed small-scale yet high-quality dataset, DriveVer can accurately evaluate the planning results of the base planner across different scenarios and guide the trajectory optimization direction, while exhibiting enhanced universality and generalization.

\begin{figure*}
    \centering
    \includegraphics[width=1.0\linewidth]{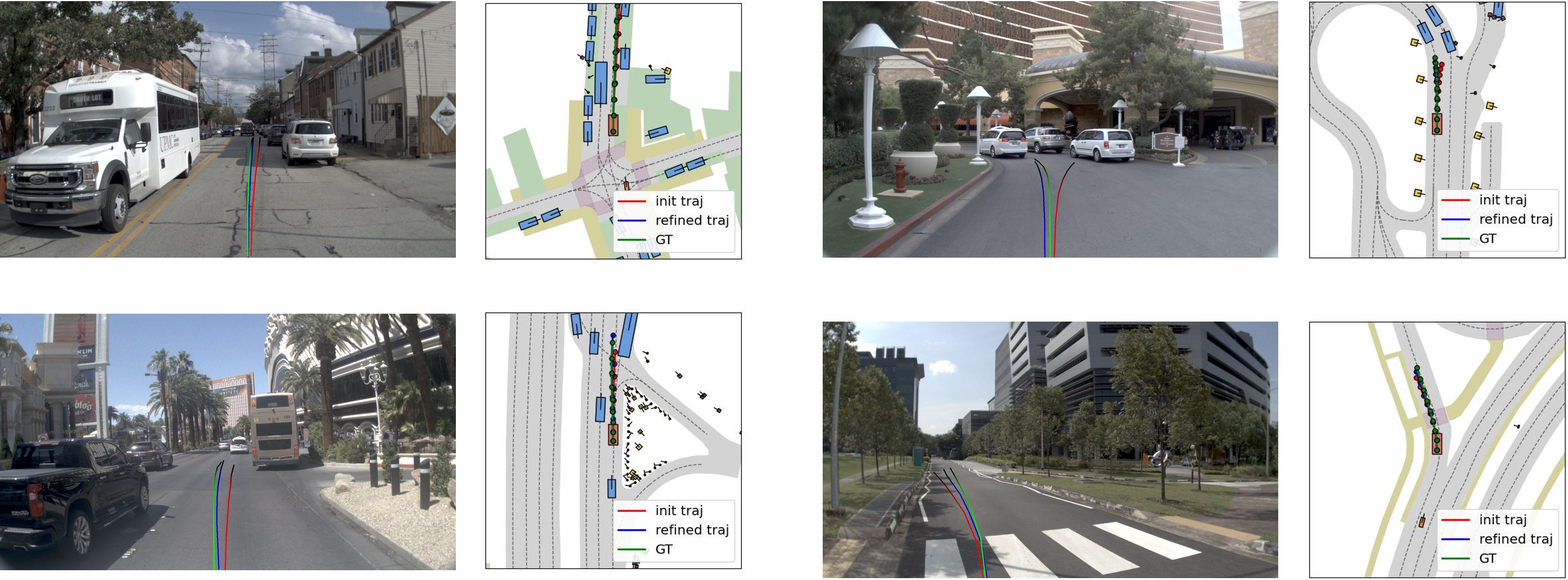}
    \caption{Qualitative visualization comparing the \textbf{initial baseline trajectory from DiffusionDrive (Red)}, the \textbf{trajectory refined by DriveVer (Blue)}, and the \textbf{Ground Truth Trajectory (Green)}.}
    \label{fig:visual}
\end{figure*}

\subsubsection{Qualitative Analysis}
Figure~\ref{fig:visual} intuitively illustrates the trajectory optimization results of DriveVer by visualizing the initial trajectory (planned by DiffusionDrive) and the refined trajectory.

In the top-left case, the initial trajectory is close to the parked vehicles on the right side of the road, posing a collision risk. DriveVer adjusts the trajectory to the center of the lane, which is consistent with the human expert trajectory. In the bottom-left case, a similar scenario is observed. The trajectory planned by DiffusionDrive is excessively close to the preceding bus, leading to a collision between the ego vehicle and the preceding vehicle. The refined trajectory generated by DriveVer increases the lateral distance to the forward bus, thus avoiding collisions and enhancing driving safety. In the top-right case, DiffusionDrive produces an unreasonable planning result, whereas DriveVer can make substantial corrections leveraging its scene-understanding capability. Similarly, in the bottom-right case, the initial trajectory deviates from the drivable area, and DriveVer can accordingly correct it to keep the ego vehicle within the valid road region.

These results further demonstrate that DriveVer has a strong understanding of desirable driving behaviors in driving scenarios: it can not only correct dangerous behaviors in the initial planning results but also reduce potential accident risk and maintain a higher safety level.

\subsubsection{Inference Speed}
DriveVer is a lightweight model with only 34M parameters, enabling efficient test-time scaling. Experimental results on a single NVIDIA 4090 GPU show that DriveVer takes only 80 $ms$ to refine a single trajectory, which meets the real-time requirements of trajectory planning tasks. It can be plugged into other end-to-end methods without introducing excessive overhead.
\subsection{Ablation Study}
\subsubsection{Effectiveness of the Confidence Branch}

\begin{table}[t!]
\setlength{\tabcolsep}{2.3pt}
\renewcommand{\arraystretch}{1.2} 
\begin{center}
\caption{Ablation of the confidence branch in PDMS.}
\label{table:ablation1}
\begin{tabular}{c|cccc} 
\toprule
Method & DiffusionDrive & DrivoR & AdaThinkDrive & ELF-VLA  \\
\midrule
  w/o confidence branch & 88.9 & 93.5 & 90.8 & 91.4 \\
\midrule
\rowcolor{gray!30}  w/ confidence branch & \textbf{89.0} & \textbf{93.8} & \textbf{90.9} & \textbf{91.5} \\
\bottomrule
\end{tabular}
\vspace{-1.5em}
\end{center}
\end{table}

As demonstrated in Tab.~\ref{table:ablation1}, the confidence branch plays a crucial role in ensuring DriveVer's overall trajectory refinement performance, particularly when the base planner has already achieved high performance. For instance, if the initial trajectories generated by DrivoR are refined indiscriminately without prior evaluation, the PDMS of the refined trajectories degrades rather than improves. This phenomenon arises because the refinement branch exhibits inevitable prediction variance when estimating small refinements for sufficiently high-quality initial trajectories. Refining these trajectories using single-inference results tends to undermine the inherent advantages of the original trajectories. The confidence branch enables DriveVer to evaluate initial trajectories and retain high-quality ones in a simple yet effective manner, while avoiding excessive test-time overhead for only marginal performance gains on such trajectories.

\subsubsection{Refinement Method}

We compare the performance of our proposed direction vector prediction with a baseline approach that directly predicts the absolute coordinate residuals. Specifically, the baseline predicts the residual of each waypoint's location $\delta \tau = \tau_{\text{gt}} - \tau_{\text{init}}$ and is supervised by an $L_1$ loss, formulated as $\mathcal{L}_{L_1} = \mathbb{E} \left[\|\hat{\delta\tau} - \delta \tau_{gt}\|_1 \right]$. In contrast, our method predicts the geometric refinement direction vector $\hat{u}$ and is supervised by a cosine similarity loss $\mathcal{L}_{\text{cos}} = \mathbb{E} \left[1-\langle \hat{u}, u_{\text{gt}} \rangle \right]$. As shown in Tab.~\ref{table:ablation2}, our directional refinement approach with the cosine loss outperforms the absolute residual baseline. Optimizing the cosine similarity guides the model to focus on the overall optimization trend of the trajectory, rather than enforcing a strict penalty on the absolute refinement magnitude of each individual waypoint. This formulation effectively reduces the harsh penalty induced by the $L_1$ loss, which helps mitigate overfitting, improves model robustness, and enables better generalization to provide reliable optimization directions in unseen scenarios.

\begin{table}[t!]
\setlength{\tabcolsep}{8pt}
\renewcommand{\arraystretch}{1.2} 
\begin{center}
\caption{Comparison of Refinement Methods}
\label{table:ablation2}
\begin{tabular}{l|l|c} 
\toprule
Refinement Target & Supervision Loss & PDMS $\uparrow$ \\
\midrule
Absolute Residual & $L_1$ Loss & 88.3 \\
\rowcolor{gray!30} Direction Vector (Ours) & Cosine Loss & \textbf{89.0} \\
\bottomrule
\end{tabular}
\end{center}
\vspace{-1.5em}
\end{table}

\section{Conclusion}

End-to-end autonomous driving models have traditionally relied heavily on training-time scaling, which incurs prohibitive computational costs and exhibits sharply diminishing returns. Moreover, most existing planners follow a one-shot trajectory generation paradigm, lacking the ability to post-hoc validate or correct suboptimal or potentially unsafe trajectories during inference.
To address these fundamental limitations, we introduce DriveVer, a lightweight, plug-and-play test-time verifier. Unlike conventional pipelines, DriveVer leverages a test-time scaling mechanism to provide post-hoc evaluation and active geometric correction for autonomous driving systems. By efficiently fusing multi-view visual representations, ego-vehicle kinematic features, and initial candidate trajectories through a cross-modal architecture, DriveVer simultaneously predicts a trajectory safety confidence score and an absolute geometric refinement vector.

We also construct a dedicated, condition-driven trajectory evaluation dataset based on the NAVSIM benchmark. DriveVer significantly tightens the safety upper bound of existing base planners, consistently improving performance across diverse baselines. Notably, as an extremely compact model with only 34$M$ parameters, DriveVer incurs minimal overhead and achieves an inference latency of just 80 $ms$, making it well-suited for real-time deployment.

\bibliographystyle{IEEEtran}
\bibliography{main}

\end{document}